\documentclass[11pt,a4paper]{article}
\usepackage[hyperref]{naaclhlt2019}
\usepackage{times}
\usepackage{latexsym}

\usepackage{url}

\usepackage{graphicx}
\usepackage{mathtools}
\usepackage{subfigure}
\usepackage{multirow}

\aclfinalcopy 
\def\aclpaperid{1182} 

\title{Improving Grammatical Error Correction via Pre-Training a Copy-Augmented Architecture with Unlabeled Data}

\author{Wei Zhao, Liang Wang, Kewei Shen, Ruoyu Jia, Jingming Liu \\
  Yuanfudao Research / Beijing, China \\
  {\tt \{zhaowei01,wangliang01,shenkw,jiary,liujm\}@fenbi.com}}

\date{}

\begin{document}
\maketitle
\begin{abstract}
Neural machine translation systems have become state-of-the-art approaches for Grammatical Error Correction (GEC) task. In this paper, we propose a copy-augmented architecture for the GEC task by copying the unchanged words from the source sentence to the target sentence. Since the GEC suffers from not having enough labeled training data to achieve high accuracy. We pre-train the copy-augmented architecture with a denoising auto-encoder using the unlabeled One Billion Benchmark and make comparisons between the fully pre-trained model and a partially pre-trained model. It is the first time copying words from the source context and fully pre-training a sequence to sequence model are experimented on the GEC task. Moreover, We add token-level and sentence-level multi-task learning for the GEC task. The evaluation results on the CoNLL-2014 test set show that our approach outperforms all recently published state-of-the-art results by a large margin. The code and pre-trained models are released at https://github.com/zhawe01/fairseq-gec.
\end{abstract}

\section{Introduction}

Grammatical Error Correction (GEC) is a task of detecting and correcting grammatical errors in text. Due to the growing number of language learners of English, there has been increasing attention to the English GEC, in the past decade.

The following sentence is an example of the GEC task, where the word in bold needs to be corrected to its adverb form.

\begin{center}
\textit{Nothing is $\left[ \textbf{absolute} \rightarrow \text{absolutely} \right]$ right or wrong.}
\end{center}

Although machine translation systems have become state-of-the-art approaches for GEC, GEC is different from translation since it only changes several words of the source sentence. In Table \ref{copy_per}, we list the ratio of unchanged words of the target sentence to the source sentence in three different datasets. We can observe that more than 80\% of the words can be copied from the source sentence.

\begin{table}[t!]
\begin{center}
\begin{tabular}{|l|r|c|l|}
\hline \bf Corpus & \bf Sent. & \bf Tok. & \bf Same \% \\ \hline
CoNLL-2013 & 1,381 & 28,944 & 96.50\% \\
JFLEG & 754 & 14,240 & 84.23\% \\
Lang-8 & 4,936 & 73,705 & 83.22\% \\
\hline
\end{tabular}
\end{center}
\caption{\label{copy_per} The ratio of unchanged words in the target sentence to the source sentence. ``Sent.'' means the sentence number. ``Tok.'' means the token number of the target sentence. ``Same \%'' means the same word percentage.}
\end{table}

Considering the percentage of unchanged words is high in the GEC task, a more proper neural architecture is needed for it. We enhance the current neural architecture by enabling it to copy the unchanged words and the out-of-vocabulary words directly from the source sentence, just as what humans do when they correct sentences. To our knowledge, this is the first time that neural copying mechanism is used on GEC.

Progresses have been made thanks to large-scale training corpus, including NUS Corpus of Learner English (NUCLE) \cite{dahlmeier2013building} and the large-scale Lang-8 corpus\cite{tajiri2012tense}. However, even with millions of labeled sentences, automatic GEC is challenging due to the lack of enough labeled training data to achieve high accuracy. 

To alleviate the problem of insufficient labeled data, we propose a method to leverage the unlabeled data. The concrete way is to pre-train our copy-augmented model with the unlabeled One Billion Benchmark \cite{chelba2013one} by leveraging denoising auto-encoders. 

We also add two multi-tasks for the copy-augmented architecture, including a token-level labeling task and a sentence-level copying task, to further improve the performance of the GEC task.

The copying mechanism is for the first time used on the GEC task, which was used on text summarization tasks. On the GEC task, copying mechanism enables training a model with a small vocabulary since it can straightly copy the unchanged and out-of-vocabulary words from the source input tokens. Besides, by separating the constant part of the work from the GEC task, copying makes the generating portion of the architecture more powerful. In the experiment section of this paper, we show that copying does more than just solving the ``UNK problem'', and it can also recall more edits for the GEC problem. 

The copy-augmented architecture outperforms all the other architectures on the GEC task, by achieving a 56.42 $F_{0.5}$ score on the CoNLL 2014 test data set. Combined with denoising auto-encoders and multi-tasks, our architecture achieves 61.15 $F_{0.5}$ on the CoNLL-2014 test data set, improving +4.9 $F_{0.5}$ score than state-of-the-art systems. 

In summary, our main contributions are as follows. (1) We propose a more proper neural architecture for the GEC problem, which enables copying the unchanged words and out-of-vocabulary words directly from the source input tokens. (2) We pre-train the copy-augmented model with large-scale unlabeled data using denoising auto-encoders, alleviating the problem of the insufficient labeled training corpus. (3) We evaluate the architecture on the CoNLL-2014 test set, which shows that our approach outperforms all recently published state-of-the-art approaches by a large margin.

\section{Our Approach}

\subsection{Base Architecture}
Neural machine translation systems have become the state-of-the-art approaches for Grammatical Error Correction (GEC), by treating the sentence written by the second language learners as the source sentence and the grammatically corrected one as the target sentence. Translation models learn the mapping from the source sentence to the target sentence. 

We use the attention based Transformer \cite{vaswani2017attention} architecture as our baseline. The Transformer encodes the source sentence with a stack of L identical blocks, and each of them applies a multi-head self-attention over the source tokens followed by position-wise feedforward layers to produce its context-aware hidden state. The decoder has the same architecture as the encoder, stacking L identical blocks of multi-head attention with feed-forward networks for the target hidden states. However, the decoder block has an extra attention layer over the encoder's hidden states.

The goal is to predict the next word indexed by t in a sequence of word tokens ($y_1, ..., y_T$), given the source word tokens ($x_1, ..., x_N$), as follows:

\begin{equation}
h^{src}_{1...N} = encoder(L^{src} x_{1...N}) \\
\end{equation}

\vspace{-3pt}
\begin{equation}
h_t=decoder(L^{trg} y_{t-1...1}, h^{src}_{1...N}) \\
\end{equation}

\vspace{-3pt}
\begin{equation}
P_t(w)=softmax(L^{trg} h_t) \\
\end{equation}

The matrix $L\in R^{d_x \times |V|}$ is the word embedding matrix, where $d_x$ is the word embedding dimension and $|V|$ is the size of the vocabulary. $h^{src}_{1...N}$ is the encoder's hidden states and $h_t$ is the target hidden state for the next word. Applying softmax operation on the inner product between the target hidden state and the embedding matrix, we get the generation probability distribution of the next word. 

\begin{equation}
l_{ce} = -\sum_{t=1}^{T}log(p_t(y_t))
\label{eqn:loss}
\end{equation}

The loss $l_{ce}$ of each training example is an accumulation of the cross-entropy loss of each position during decoding.

\subsection{Copying Mechanism}
Copying mechanism was proved effective on text summarization tasks \cite{see2017get,gu2016incorporating} and semantic parsing tasks \cite{jia2016data}. In this paper, we apply the copying mechanism on GEC task, for the first time, enabling the model to copy tokens from the source sentence.

\begin{figure*}
\begin{center}
\includegraphics[width=0.85\linewidth]{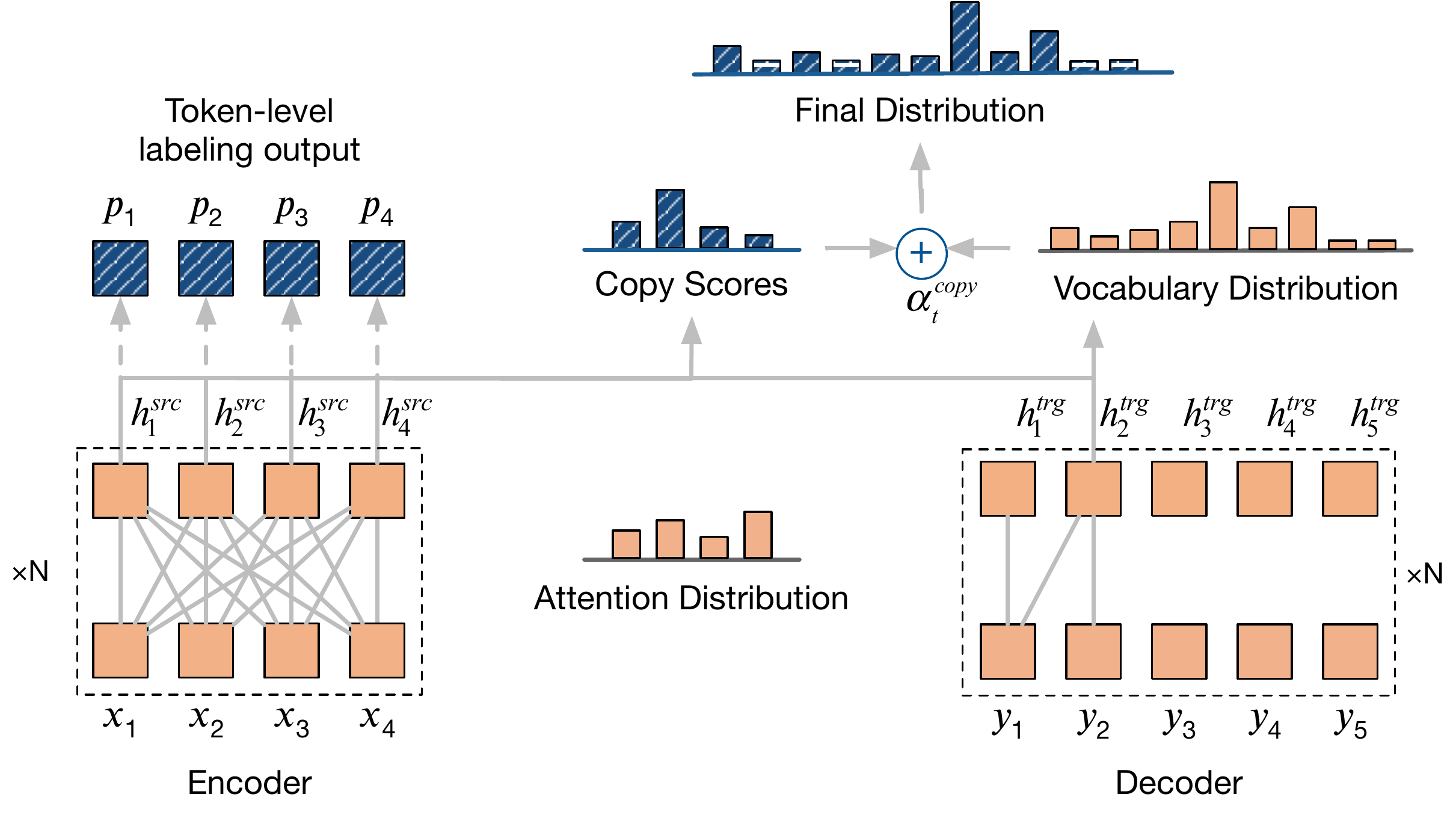}
\caption{Copy-Augmented Architecture.}
\label{fig:arch}
\end{center}
\end{figure*}

As illustrated in Figure \ref{fig:arch}, besides generating words from a fixed vocabulary, our copy-augmented network allows copying words from the source input tokens.  Defined in Equation \ref{eqn:final_distribution}, the final probability distribution $P_t$ is a mix of the generation distribution $P_t^{gen}$ and the copy distribution $P_t^{copy}$. As a result, the fixed vocabulary is extended by all the words appearing in the source sentence. The balance between the copying and generating is controlled by a balancing factor $\alpha^{copy}_t \in [0, 1]$ at each time step t.

\begin{equation}
p_t(w)=(1-\alpha^{copy}_t)*p_t^{gen}(w) + (\alpha^{copy}_t)*p^{copy}_t(w)
\label{eqn:final_distribution}
\end{equation}

The new architecture outputs the generation probability distribution as the base model, by generating the target hidden state. The copying score over the source input tokens is calculated with a new attention distribution between the decoder's current hidden state $h^{trg}$ and the encoder's hidden states $H^{src}$ (same as $h^{src}_{1...N}$). The copy attention is calculated the same as the encoder-decoder attentions, listed in Equation \ref{eqn:qkv}, \ref{eqn:attn}, \ref{eqn:softmax} :

\begin{equation}
q_t, K, V=h^{trg}_t W_q^T, H^{src} W_k^T, H^{src} W_v^T
\label{eqn:qkv}
\end{equation}

\vspace{-3pt}
\begin{equation}
A_t=q_t^T K
\label{eqn:attn}
\end{equation}

\vspace{-3pt}
\begin{equation}
P^{copy}_t(w)=softmax(A_t)
\label{eqn:softmax}
\end{equation}

The $q_t$, $K$ and $V$ are the query, key, and value that needed to calculate the attention distribution and the copy hidden state. We use the normalized attention distribution as the copy scores and use the copy hidden states to estimate the balancing factor $\alpha_t^{copy}$. 

\begin{equation}
\alpha^{copy}_t=sigmoid(W^T \sum(A^T_t \cdot V))
\end{equation}

The loss function is as described in Equation \ref{eqn:loss}, but with respect to our mixed probability distribution $y_t$ given in Equation \ref{eqn:final_distribution}.

\section{Pre-training}
Pre-training is shown to be useful in many tasks when lacking vast amounts of training data. In this section, we propose denoising auto-encoders, which enables pre-training our models with large-scale unlabeled corpus. We also introduce a partially pre-training method to make a comparison with the denoising auto-encoder. 

\subsection{Denoising Auto-encoder}
Denoising auto-encoders \cite{vincent2008extracting} are commonly used for model initialization to extract and select features from inputs. BERT \cite{devlin2018bert} used a pre-trained bi-directional transformer model and outperformed existing systems by a wide margin on many NLP tasks. In contrast to denoising auto-encoders, BERT only predicts the 15\% masked words rather than reconstructing the entire input. BERT denoise the 15\% of the tokens at random by replacing 80\% of them with [MASK], 10\% of them with a random word and 10\% of them unchanged. 

Inspired by BERT and denoising auto-encoders, we pre-traine our copy-augmented sequence to sequence model by noising the One Billion Word Benchmark \cite{chelba2013one}, which is a large sentence-level English corpus. In our experiments, the corrupted sentence pairs are generated by the following procedures.

\begin{itemize}
\item Delete a token with a probability of 10\%.
\item Add a token with a probability of 10\%. 
\item Replace a word with a randomly picked word from the vocabulary with a probability of 10\%.
\item Shuffle the words by adding a normal distribution bias to the positions of the words and re-sort the words by the rectified positions with a standard deviation 0.5.
\end{itemize}

With a large amount of the artificial training data, the sequence to sequence model learns to reconstruct the input sentence, by trusting most of the input tokens but not always. A sentence pair generated by the corruption process is a GEC sentence pair to some degree, since both of them are translating a not ``perfect'' sentence to a ``perfect'' sentence by deleting, adding, replacing or shuffling some tokens.

\subsection{Pre-training Decoder}
In nature language processing (NLP), pre-training part of the model also improves many tasks' performance. Word2Vec and GloVe \cite{pennington2014glove,mikolov2013distributed} pre-trained word embeddings. CoVe \cite{mccann2017learned} pre-trained a encoder. ELMo \cite{peters2018deep} pre-trained a deep bidirectional architecture, and etc. All of them are shown to be effective in many NLP tasks.

Following \cite{ramachandran2016unsupervised,junczys2018approaching}, we experiment with pre-training the decoder of the copy-augmented sequence-to-sequence architecture as a typical language model. We initialize the decoder of the GEC model with the pre-trained parameters, while initializing the other parameters randomly. Since we use the tied word embeddings between encoder and decoder, most parameters of the model are pre-trained, except for those of the encoder, the encoder-decoder's attention and the copy attention.

\section{Multi-Task Learning}

The Multi-Task Learning (MTL) solves problems by jointly training multiple related tasks, and has shown its advantages in many tasks, ranging from computer vision \cite{zhang2014facial,dai2016instance} to NLP \cite{collobert2008unified,sogaard2016deep}. In this paper, we explore two different tasks for GEC to improve the performance.

\subsection{Token-level Labeling Task}
We propose a token-level labeling task for the source sentence, and assign each token in the source sentence a label indicating whether this token is right/wrong. 

Assuming that each source token $x_i$ can be aligned with a target token $y_j$, we define that the source token is right if $x_i=y_j$, and wrong otherwise. Each token's label is predicted by passing the final state $h^{src}_i$ of the encoder through a softmax after an affine transformation, as shown in Equation \ref{eqn:label_task}.

\begin{equation}
p(label_i|x_{1...N}) = softmax(W^Th^{src}_i)
\label{eqn:label_task}
\end{equation}

This token-level labeling task explicitly augment the input tokens' correctness to the encoder, which can later be used by the decoder.

\subsection{Sentence-level Copying Task}

The primary motivation behind the sentence-level copying task is to make the model do more copying when the input sentence looks entirely correct. 

During training, we send equal number of sampled correct sentence pairs and the edited sentence pairs to the model. When inputting the right sentences, we remove the decoder's attention over the outputs of the encoder. Without the encoder-decoder attention, the generating work gets hard. As a result, the copying part of the model will be boosted for the correct sentences.

\section{Evaluations}

\subsection{Datasets}
As previous studies, we use the public NUCLE \cite{dahlmeier2013building}, Lang-8 \cite{tajiri2012tense} and FCE \cite{yannakoudakis2011new} corpus as our parrallel training data. 
The unlabeled dataset we use is the well-known One Billion Word Benchmark \cite{chelba2013one}.
We choose the test set of CoNLL-2014 shared task as our test set and CoNLL-2013 test data set \cite{dahlmeier2013building} as our development benchmark. For the CoNLL data sets, the MaxMatch ($M^2$) scores \cite{dahlmeier2012better} were reported, and for the JFLEG \cite{napoles2017jfleg} test set, the GLEU metric \cite{sakaguchi2016reassessing} were reported.

To make our results comparable to state-of-the-art results in the field of GEC, we limit our training data strictly to public resources. Table \ref{train_data} and Table \ref{test_data} list all the data sets that we use in this paper.

\begin{table}[t!]
\begin{center}
\begin{tabular}{|l|r|l|l|}
\hline \bf Corpus & \bf Sent. & \bf Public & \bf Type \\ \hline
Lang-8 & 1,097,274 & Yes & Labeled \\
NUCLE & 57,119 & Yes & Labeled \\
FCE & 32,073 & Yes & Labeled \\
\hline
One-Billion & 30,178,573 & Yes & Unlabeled \\
\hline
\end{tabular}
\end{center}
\caption{\label{train_data} Training Corpus}
\end{table}

\begin{table}[t!]
\begin{center}
\begin{tabular}{|l|r|c|l|}
\hline \bf Corpus & \bf Sent. & \bf Annot. & \bf Metric \\ \hline
CoNLL-2013 & 1,381 & 1 & $M^2$ \\
CoNLL-2014 & 1,312 & 2 & $M^2$ \\
JFLEG & 747 & 4 & GLEU \\
\hline
\end{tabular}
\end{center}
\caption{\label{test_data} Evaluation Corpus}
\end{table}

We build a statistical-based spell error correction system and correct the spell errors in our training data. Following \cite{ge2018reaching,junczys2018approaching,chollampatt2018multilayer} and etc., we apply spell correction before evaluation for our dev/test datasets. A 50,000-word dictionary is extracted from the spell-corrected Lang-8 data corpus. Like previous works, we remove the unchanged sentence pairs in the Lang-8 corpus before training.

\subsection{Model and Training Settings}
In this paper, we use the Transformer implementation in the public FAIR Sequence-to-Sequence Toolkit \footnote{https://github.com/pytorch/fairseq} \cite{gehring2017convolutional} codebase. 

For the transformer model, we use token embeddings and hidden size of dimension 512, and the encoder and decoder have 6 layers and 8 attention heads. For the inner layer in the positionwise feed-forward network, we use 4096. Similar to previous models we set the dropout to 0.2. A 50,000 vocabulary for the input and output tokens are collected from the training data. In total, this model has 97M parameters.

Models are optimized with Nesterov’s Accelerated Gradient \cite{nesterov1983method}. We set the learning rate with 0.002, the weight decay 0.5, the patience 0, the momentum 0.99 and minimum learning rate 10-4. During training, we evaluate the performance on the development set for every epoch.

We also use edit-weighted MLE objective as \cite{junczys2018approaching}, by scaling the loss of the changed words with a balancing factor $\Lambda$.

Almost the same architecture and hyper-parameters are used when pre-training using unlabeled data, except the $\Lambda$ parameter for edit-weighted loss. We set $\Lambda=3$ when we train the denoising auto-encoder, and set $\Lambda \in [1, 1.8]$ when we train GEC models. 

During decoding, we use a beam-size of 12 and normalize model scores by length. We do not use reranking when evaluating the CoNLL-2014 data sets. But we rerank the top 12 hypothesizes using the language model trained on Common Crawl \cite{junczys2016phrase} for the JFLEG test sets.

\subsection{Experimental Results}
We compare our results with the well-known GEC systems, as shown in Table \ref{comparison}. Rule, classification, statistical machine translation (SMT), and neural machine translation (NMT) based systems were built for the GEC task. We list the well-known models on the top section of Table \ref{comparison} and our results in the middle. Almost all the previous systems reranked their top 12 results using a big language model and some of them used partially pre-trained parameters, which improve their results by 1.5 to 5 $F_{0.5}$ score.  Our copy-augmented architecture achieve a 56.42 $F_{0.5}$ score on the CoNLL-2014 dataset and outperforms all the previous architectures even without reranking or pre-training. 

Combined with denoising auto-encoders and multi-tasks, our model achieve a 61.15 $F_{0.5}$ score on the CoNLL-2014 data set. This result exceeds the previous state-of-the-art system +4.9 $F_{0.5}$ points.

In the bottom section of Table \ref{comparison}, we list the results of \cite{ge2018reaching}. No direct comparison can be made between us, because they used the non-public Cambridge Learner Corpus (CLC) \cite{nicholls2003cambridge} and their own collected non-public Lang-8 corpus, making their labeled training data set 3.6 times larger than ours. Even so, our results on the CoNLL 2014 test data set and JFLEG test data set are very close to theirs.

\begin{table*}[t!]
\begin{center}
\begin{tabular}{|l|c|ccc|c|c|}
\hline 
\bf \multirow{2}{*}{Model} & \bf \multirow{2}{*}{Year} & \multicolumn{3}{|c|}{\bf CoNLL-14} & \bf JFLEG & \bf \multirow{2}{*}{Dict} \\
\bf & \bf & \bf Pre. & \bf Rec. & \bf $F_{0.5}$ & \bf GLEU & \bf  \\ \hline
SMT (with LM) & 2014 & 41.72 & 22.00 & 35.38 & - & word \\
SMT Rule-Based Hybird (with LM) & 2014 & 39.71 & 30.10 & 37.33 & - & word \\
SMT Classification Hybird (with LM) & 2016 & 60.17 & 25.64 & 47.40 & - & word \\
Neural Hybird MT (with LM) & 2017 & - & - & 45.15 & 53.41 & char/word \\
CNN + EO (4 ens. with LM) & 2018 & 65.49 & 33.14 & 54.79 & 57.47 & bpe \\
Transformer + MIMs (4 ens. with LM) & 2018 & 63.00 & 38.90 & 56.10 & 59.90 & bpe \\
NMT SMT Hybrid (4 ens. with LM) & 2018 & 66.77 & 34.49 & 56.25 & 61.50 & bpe \\

\hline 
\multicolumn{7}{|l|}{\bf Our Model} \\
\hline
Copy-augmented Model (4 ens.) & - & 68.48 & 33.10 & \bf 56.42 & $59.48^{*}$ & word \\
+ DA, Multi-tasks (4 ens.) & - & 71.57 & 38.65 & \bf 61.15 & $61.00^{*}$ & word \\

\hline
\multicolumn{7}{|l|}{\bf Model Trained with Large Non-public Training Data} \\
\hline
CNN + FB Learning (4 ens. with LM) & 2018 & 74.12 & 36.30 & 61.34 & 61.41 & bpe \\

\hline
\end{tabular}
\end{center}
\caption{\label{comparison} Comparison of GEC systems on CoNLL-2014 and JFLEG test set. The $M^2$ score for CoNLL-2014 test dataset and the GLEU for the JFLEG test set are reported. DA refers to the "Denoising Auto-encoder". (with LM) refers to the usage of an extra language model. (4 ens.) refers to the ensemble decoding of 4 independently trained models. We re-rank the results of the top 12 hypothesizes for the JFLEG test set with an extra language model and marked them with $^*$.}
\end{table*}

In Table \ref{comparison}, ``SMT (with LM)'' refers to \cite{junczys2014amu}; ``SMT Rule-Based Hybird'' refers to \cite{felice2014grammatical}; ``SMT Classification Hybird'' refers to \cite{rozovskaya2016grammatical}; ``Neural Hybird MT'' refers to \cite{ji2017nested}; ``CNN + EO'' refers to \cite{chollampatt2018multilayer} and ``EO'' means rerank with edit-operation features; ``Transformer + MIMs'' refers to \cite{junczys2018approaching} and ``MIMs'' means model indepent methods; ``NMT SMT Hybrid'' refers to \cite{grundkiewicz2018near}; ``CNN + FB Learning'' refers to \cite{ge2018reaching}.

\subsection{Ablation Study}

\subsubsection{Copying Ablation Results}

In this section, we compare the Transformer architecture's results with and without copying mechanism on the GEC task. As illustrated in Table \ref{ablation}, copy-augmented model increases the $F_{0.5}$ score from 48.07 to 54.67, with a +6.6 absolute increase. Most of the improvements come from the words that are out of the fixed vocabulary, which will be predicted as a UNK word in the base model but will be copied as the word itself in the copy-augmented model. 

Copying is generally known as good at handling the UNK words. To verify if copying is more than copying UNK words, we do experiments by ignoring all UNK edits. From Table \ref{ablation}, we can see that even ignoring the UNK benefits, the copy-augmented model is still 1.62 $F_{0.5}$ points higher than the baseline model, and most of the benefit comes from the increased recall.

\begin{table*}[t!]
\begin{center}
\begin{tabular}{|l|r|r|r|c|}
\hline \bf Model & \bf Pre. & \bf Rec.  & \bf $F_{0.5}$ & \bf Imp. \\ \hline
Transformer & 55.96 & 30.73 & 48.07 & - \\
+ Copying & 65.23 & 33.18 & \bf 54.67 & +6.60 \\

\hline \multicolumn{5}{|l|}{\bf Ignoring UNK words as edits} \\ \hline
Transformer & 65.26 & 30.63 & 53.23 & - \\
+ Copying & 65.54 & 33.18 & 54.85 & +1.62 \\

\hline \multicolumn{5}{|l|}{\bf + Pre-training} \\ \hline
Copy-Augmented Transformer & 65.23 & 33.18 & 54.67 & - \\
+ Pre-training Decoder (partially pre-trained) & 68.02 & 34.98 & 57.21 & +2.54  \\
+ Denosing Auto-encoder (fully pre-trained)& 68.97 & 36.98 & \bf 58.80 & +4.13 \\

\hline \multicolumn{5}{|l|}{\bf + Multi-tasks} \\ \hline
Copy-Augmented Transformer & 67.74 & 40.62 & \bf 59.76 & - \\

\hline
\end{tabular}
\end{center}
\caption{\label{ablation} Single Model Ablation Study on CoNLL 2014 Test Data Set.}
\end{table*}

\subsubsection{Pre-training Ablation Results}
From Table \ref{ablation}, we can observe that by partially pre-training the decoder, the $F_{0.5}$ score is improved from 54.67 to 57.21 (+2.54). It is an evident improvment compared to the un-pre-trained ones. However, the denoising auto-encoder improves the single model from 54.67 to 58.8 (+4.13). We can also see that both the precision and recall are improved after pre-training.

To further investigate how good the pre-trained parameters are, we show the results of the early stage with and without the denoising auto-encoder's pre-trained parameters in Table \ref{finetune}. The results show, if we finetune the model for 1 epoch with the labeled training data, the pre-trained model beats the un-pretrained one with a big gap (48.89 vs 17.19). Even without finetune, the pre-trained model can get a $F_{0.5}$ score of 31.33. This proves that pre-training gives the models much better initial parameters than the randomly picked ones.

\begin{table}[t!]
\begin{center}
\begin{tabular} {|l|r|r|r|}
\hline \bf Finetune & \bf Pre. & \bf Rec. & \bf $F_{0.5}$ \\ 
\hline \multicolumn{4}{|l|}{\bf with the denoising auto-encoder} \\ \hline
no finetune & 36.61 & 19.87 & 31.33 \\
finetune 1 epoch & 68.58 & 22.76 & 48.89 \\
\hline \multicolumn{4}{|l|}{\bf without the denoising auto-encoder} \\ \hline
finetune 1 epoch & 32.55 & 05.96 & 17.19 \\
\hline
\end{tabular}
\end{center}
\caption{\label{finetune} Denoising Auto-encoder's Results on CoNLL-2014 Test Data Set.}
\end{table}

\subsubsection{Sentence-level Copying Task Ablation Results} 

We add the sentence-level copying task to encourage the model outputs no edits when we input a correct sentence. To verify this, we create a correct sentence set by sampling 500 sentences from Wikipedia. Also, we generate an error sentence set by sampling 500 sentences from CoNLL-2013 test data set, which is an error-annotated dataset. Then we calculate the average value of the balance factor $\alpha^{copy}$ of the two sets. 

Before we add the sentence-level copying task, the $\alpha^{copy}$ is 0.44/0.45 for the correct and error sentence sets. After adding the sentence-level copying task, the value changed to 0.81/0.57. This means that 81\% of the final score comes from copying on the correct sentence set, while only 57\% on the error sentence set. By adding the sentence-level copying task, models learn to distinguish correct sentences and error sentences.

\subsection{Attention Visualization}

\begin{figure*}
\begin{center}
\subfigure[Copy Alignment]{
\includegraphics[width=\textwidth]{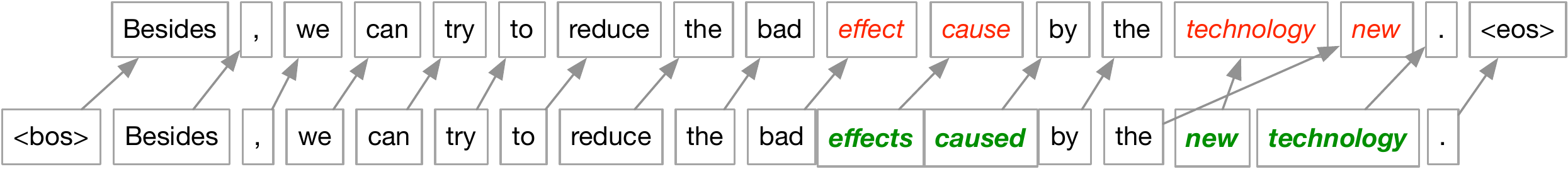}
\label{fig:copy}}

\subfigure[Encoder-Decoder Atttention Alignment]{
\includegraphics[width=\textwidth]{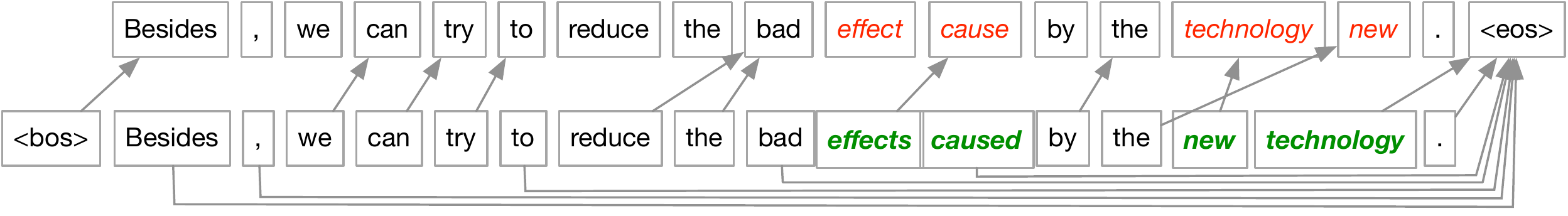}
\label{fig:gen}}

\caption{An example of the different behaviors between the copy and encoder-decoder attention. In each figure, the above line is the source sentence, where the error words are in italic. The bottom line is the corrected sentence, where the corrected words are in bold italic. The arrow means which source token the copy and encoder-decoder attention mainly focus on, when predicting the current word. ``$\langle bos \rangle$'' refers to the begin of the sentence and ``$\langle eos \rangle$'' refers to the end of the sentence.}
\label{fig:align}
\end{center}
\end{figure*}

To analyze how copying and generating divide their work. We visualized the copying attention alignment and the encoder-decoder attention alignment in Figure \ref{fig:align}. In Figure \ref{fig:copy}, copying focus their weights on the next word in good order, while in Figure \ref{fig:gen}, generating moves its attention more on the other words, e.g., the nearby words, and the end of the sentence. As explained in \cite{raganato2018analysis}, this means that the generating part tries to find long dependencies and attend more on global information.

By separating the copying work from the generation work, the generation part of the model can focus more on the ``creative'' works.

\section{Discussion}

\subsection{Recall on Different Error Types}
Automatic grammatical error correction is a complicated task since there are different kinds of errors and various correction ways. In this section, we analyze our systems' performance on different grammatical error types. \cite{ng2014conll} labeled CoNLL-2014 test set with 28 error types, and we list the recall percentage on the top 9 error types. We summarize the other 19 types in the last line of the table.

\begin{table}[t!]
\begin{center}
\begin{tabular}{|l|r|r|r|}
\hline \bf Error Type & \bf \% & \bf Recall \\ \hline
Article Or Determiner & 14.31\% & 44.54\% \\
Wrong Collocation/Idiom & 12.75\% & \bf{10.38\%} \\
Spelling, Punctuation, etc. & 12.47\% & 45.66\% \\
Preposition & 10.38\% & 49.03\% \\
Noun number & 9.38\% & \bf{72.65\%} \\
Verb Tense & 5.41\% & 28.15\% \\
Subject-Verb Agreement & 4.93\% & \bf{61.79\%} \\
Verb form & 4.69\% & 57.26\% \\
Redundancy & 4.65\% & 25.86\% \\ \hline

Others & 20.99\% & 23.28\% \\
\hline
\end{tabular}
\end{center}
\caption{\label{error_type} Recall on Different Error Types. \% is the percentage of this error type in the test data set. Recall is the percentage of the fixed errors in each error type.}
\end{table}

Our approach recalls 72.65\% errors on the ``Noun number'' type and 61.79\% on the ``Subject-Verb Agreement'' type. However, only 10.38\% errors are recalled on the ``Wrong Collocation/Idiom'' type. 

Computers are good at the definite and mechanical errors, but still have a big gap with humans on the error types that are subjective and with cultural characteristics.

\section{Related Work}

Early published works in GEC develop specific classifiers for different error types and then use them to build hybrid systems. 
Later, leveraging the progress of statistical machine translation(SMT) and large-scale error corrected data, GEC systems are further improved treated as a translation problem. SMT systems can remember phrase-based correction pairs, but they are hard to generalize beyond what was seen in training. The CoNLL-14 shared task overview paper \cite{ng2014conll} provides a comparative evaluation of approaches. \cite{rozovskaya2016grammatical} detailed classification and machine translation approaches to grammatical error correction problems, and combined the strengths for both methods.

Recently, neural machine translation approaches have been shown to be very powerful. \cite{yannakoudakis2017neural} developed a neural sequence-labeling model for error detection to calculate the probability of each token in a sentence as being correct or incorrect, and then use the error detecting model's result as a feature to re-rank the N best hypotheses. \cite{ji2017nested} proposed a hybrid neural model incorporating both the word and character-level information. \cite{chollampatt2018multilayer} used a multilayer convolutional encoder-decoder neural network and outperforms all prior neural and statistical based systems on this task. \cite{junczys2018approaching} tried deep RNN \cite{barone2017deep} and transformer \cite{vaswani2017attention} encoder-decoder models and got a higher result by using transformer and a set of model-independent methods for neural GEC. 

The state-of-the-art system on GEC task is achieved by \cite{ge2018reaching}, which are based on the sequence-to-sequence framework and fluency boost learning and inference mechanism. However, the usage of the non-public CLC corpus \cite{nicholls2003cambridge} and self-collected non-public error-corrected sentence pairs from Lang-8 made their training data 3.6 times larger than the others and their results hard to compare.

\section{Conclusions}

We present a copy-augmented architecture for GEC, by considering the characteristics of this problem. Firstly, we propose an enhanced copy-augmented architecture, which improves the sequence-to-sequence model's ability by directly copying the unchanged words and out-of-vocabulary words from the source input tokens. Secondly, we fully pre-train the copy-augmented architecture using large-scale unlabeled data, leveraging denoising auto-encoders. Thirdly, we introduce two auxiliary tasks for multi-task learning. Finally, we outperform the state-of-the-art automatic grammatical error correction system by a large margin. However, due to the complexity of the GEC problem, there is still a long way to go to make the automatic GEC systems as reliable as humans.

\bibliography{naaclhlt2019}
\bibliographystyle{acl_natbib}

\end{document}